\documentclass[twoside,11pt]{article}

\usepackage{jmlr2e}

\usepackage{url}
\usepackage[utf8]{inputenc} 
\usepackage[T1]{fontenc}    
\usepackage{booktabs}       
\usepackage{amsfonts}       
\usepackage{nicefrac}       
\usepackage{microtype}      
\usepackage[usenames,dvipsnames]{pstricks}
\usepackage{epsfig}
\usepackage{pst-grad} 
\usepackage{pst-plot} 
\usepackage{amsmath}
\usepackage{amssymb}
\usepackage{theorem}
\usepackage{mathrsfs} 
\usepackage{euscript}
\usepackage{graphicx}
\usepackage{float}
\usepackage{textcomp}
\usepackage{listings}
\usepackage{xcolor}
\usepackage{cases}
\usepackage{algorithm}
\usepackage{algorithmic}
\usepackage{multirow}
\usepackage[lofdepth,lotdepth]{subfig}

\graphicspath{{figures/}}
\definecolor{labelkey}{rgb}{0,0.08,0.45}
\definecolor{refkey}{rgb}{0,0.6,0.0}
\definecolor{Brown}{rgb}{0.45,0.0,0.05}
\definecolor{dgreen}{rgb}{0.00,0.49,0.00}
\definecolor{dblue}{rgb}{0,0.08,0.75}
\definecolor{nido}{rgb}{0.6,0.0,0.4}
\renewcommand{\leq}{\ensuremath{\leqslant}}
\renewcommand{\geq}{\ensuremath{\geqslant}}

\newcommand{\RR}{\ensuremath{\mathbb{R}}}

\newtheorem{theorem}{Theorem}

\newtheorem{problem}{Problem}
\newtheorem{lemma}{Lemma}
\newtheorem{proposition}{Proposition}
\newtheorem{corollary}{Corollary}

\newcommand{\tr}{\mathrm{tr}}

\newcommand{\sign}{\mathrm{sign}}
\newcommand{\eg}{\emph{e.g.}}
\newcommand{\veps}{\varepsilon}
\renewcommand{\epsilon}{\varepsilon}
\newcommand{\vphi}{\varphi}
\newcommand{\iy}{\infty}

\jmlrheading{1}{2018}{}{3/18}{}{Cyprien Gilet, Marie Deprez, Jean-Baptiste Caillau and Michel Barlaud}

\ShortHeadings{Clustering with feature selection using alternating minimization}{Gilet, Deprez, Caillau and Barlaud}
\firstpageno{1}

\begin{document}
\title{Clustering with feature selection using alternating minimization and a projection-gradient method}

\author{\name Cyprien Gilet \email gilet@i3s.unice.fr \\
       \addr I3S, Univ.\ C\^ote d'Azur \& CNRS, F-06900 Sophia Antipolis.
       \AND
       \name Marie Deprez \email deprez@ipmc.cnrs.fr \\
       \addr IPMC, Univ.\ C\^ote d'Azur \& CNRS, F-06560 Sophia Antipolis.
       \AND
       \name Jean-Baptiste Caillau \email jean-baptiste.caillau@univ-cotedazur.fr \\
       \addr LJAD, Univ.\ C\^ote d'Azur \& CNRS/Inria, F-06108 Nice.
       \AND
       \name Michel Barlaud \email barlaud@i3s.unice.fr \\
       \addr I3S, Univ.\ C\^ote d'Azur \& CNRS, F-06900 Sophia Antipolis.}

\editor{ }
 
\maketitle

\begin{abstract}
This paper deals with unsupervised clustering with feature selection in high dimensional space. The problem is to estimate both labels and a sparse projection
matrix of weights. To address this combinatorial
non-convex problem maintaining a strict control on the sparsity of the matrix of
weights, we propose an alternating minimization of the Frobenius norm criterion.
We provide a new efficient algorithm named k-sparse which alternates k-means
with projection-gradient minimization. The projection-gradient step is a method of
splitting type, with exact
projection on the $\ell^1$ ball to promote sparsity.
The convergence of the  gradient-projection step is addressed,
and a preliminary analysis of the alternating minimization is made.
Experiments on Single Cell RNA sequencing datasets show that our method significantly improves the
results of PCA k-means, spectral clustering, SIMLR,
and Sparcl methods. The complexity of our method is linear in the number of
samples (cells), so that the method scales up to large datasets.
\end{abstract}

\section{Introduction} \label{s1}
\noindent This paper deals with unsupervised clustering and feature selection in high dimensional space. Early work on feature selection were based on support vector machine (see \cite{Guyon_genes}) or logistic regression (\cite{shevade2003}). We advocate the use of sparsity promoting methods as they allow not only to perform feature selection (a crucial task in biological applications, \emph{e.g.} where features are genes), but also to use efficient state-of-the-art algorithms from convex optimization.
Clustering in high dimension using classical algorithms such as k-means (\cite{kmeansMcQueen,kmeanspp}) suffers from the curse
of dimensionality. As dimensions increase, vectors become indiscernible and the
predictive power of the aforementioned methods is drastically reduced (\cite{curse,hub}).
In order to overcome this issue, a popular approach for high-dimensional data
is to perform \emph{Principal Component Analysis} (PCA) prior to clustering.
This approach is however difficult to justify in general (\cite{PCAbefore}).
An alternative approach proposed in (\cite{kanade,Ding}) is to combine clustering and
dimension reduction by means of \emph{Linear Discriminant Analysis} (LDA). 
The heuristic used in (\cite{Ding}) is based on alternating minimization, which consists
in iteratively computing a projection subspace by LDA, using the labels $y$ at the current
iteration and then running k-means on the projection of the data onto the subspace. 
Departing from this work, \cite{diffrac} propose a convex relaxation in
terms of a suitable semi-definite program (SDP). 
Another efficient approach is spectral clustering where the main tools are graph Laplacian matrices (\cite{Ng2002,Spectral}).
However, methods such as PCA, LDA or, more recently SIMLR, do not provide sparsity.
A popular approach for selecting sparse features in supervised classification or
regression is the \emph{Least Absolute Shrinkage and Selection Operator} (LASSO)
formulation (\cite{tRS}). 
The LASSO formulation uses the $\ell^1$ norm instead of $\ell^0$
(\cite{candes,cwbES,doel,dlSR}) as an added penalty term. A hyperparameter, which
unfortunaltely does not have any simple interpretation, is then used to tune 
sparsity.
\cite{witten2010} use a lasso-type penalty to select the features and propose a sparse k-means method.
A main issue is that optimizing the values of the Lagrangian parameter $\lambda$ (\cite{hrtzER,witten2010}) is computationally expensive (\cite{myCA}). 
All these methods (\cite{diffrac,kanade,Ding,witten2010}) require a
k-means heuristic to retrieve the labels.
The alternating scheme we propose combines such a k-means step with dimension reduction,
as well as feature selection using an $\ell^1$ sparsity constraint.

\section{Constrained unsupervised classification} \label{s2}
\subsection{General Framework}
\noindent Let  $X$ be the (nonzero) $m \times d$ matrix made of $m$ line samples $x_1,\dots,x_m$
belonging to the $d$-dimensional space of features.
Let $Y \in \{0,1\}^{m\times k}$ be the matrix of labels where  $k \geq 2$
is the number of clusters. Note that we assume that this number is known; It is indeed the case for the applications we present in Section~\ref{s3}, while estimating $k$ is in general a delicate matter out of the scope of this paper. 
Each line of $Y$ has exactly one nonzero element equal
to one, $y_{ij}=1$ indicating that the sample $x_i$ belongs to the $j$-th cluster.
Let $W\in \mathbb{R}^{d\times\bar{d}}$ be the projection matrix, where the dimension in the projected space, $\bar{d}$, is understood to be much smaller than $d$.
Let then $\mu$ be the $k \times \bar{d}$ matrix of centroids of the projected data, $XW$:
\[ \mu(j,:) := \frac{1}{\sum_{i=1}^m y_{ij}} \sum_{i \text{ s.t. } y_{ij}=1} (XW)(i,:). \]
The $j$-th centroid
is the model for all samples $x_i$ belonging to the $j$-th cluster ($y_{ij}=1$).
The clustering criterion can be cast as the \emph{Within-Cluster Sum of Squares} (WCSS, \cite{selim-1984a,witten2010}) in the projected space 
\begin{equation} \label{cost}
  \frac{1}{2} \|Y\mu-XW\|_F^2 \to \min
\end{equation}
where $\|.\|_F$ is the Frobenius norm induced by the Euclidean structure on
$m \times \bar{d}$ matrices,
\[ (A|B)_F := \tr(A^{T} B) = \tr(AB^{T}),\quad \|A\|_F:=\sqrt{(A|A)_F}\,. \]
The matrix of labels is constrained according to
\begin{equation} \label{constr1}
  y_{ij} \in \{0,1\},\quad i=1,\dots,m,\quad j=1,\dots,k,
\end{equation}
\begin{equation} \label{constr1b}
  \sum_{j=1}^k y_{ij}=1,\quad i=1,\dots,m,
\end{equation}
\begin{equation} \label{constr2}
  \sum_{i=1}^m y_{ij} \geq 1,\quad j=1,\dots,k. 
\end{equation}
Note that (\ref{constr1b}) implies that each sample belongs to exactly one cluster
while (\ref{constr2}) ensures that each cluster is
not empty (no fusion of clusters). This prevents trivial solutions consisting in $k-1$
empty clusters and $W=0$.
In contrast with the Lagrangian LASSO formulation, we want to have a direct control
on the value of the $\ell^1$ bound, so we constrain $W$
according to
\begin{equation} \label{constr3}
  \|W\|_1 \leq \eta \quad (\eta>0),
\end{equation}
where $\|.\|_1$ is the $\ell^1$ norm of the vectorized $d \times \bar{d}$ matrix of
weights:
\[ \|W\|_1 := \|W(:)\|_1 = \sum_{i=1}^d \sum_{j=1}^{\bar{d}} |w_{ij}|. \]
The problem is to estimate labels $Y$ together with the sparse
projection matrix $W$. As $Y$ and $W$ are bounded, the set of constraints is
compact and existence of minimizers holds.

\begin{proposition}
The minimization of the norm (\ref{cost}), jointly in $Y$ and $W$ under the
constraints (\ref{constr1})-(\ref{constr3}), has a solution.
\end{proposition}


\noindent To attack this difficult nonconvex problem, we propose 
an alternating (or Gauss-Seidel) scheme as in \cite{kanade,Ding,witten2010}.
Another option would be to design a global
convex relaxation to address the joint minimization in $Y$ and $W$
(see, \eg, \cite{diffrac,Bach2016}) The first convex subproblem
is to find the best projection from dimension $d$ to dimension $\bar{d}$
for a given clustering.

\begin{problem} \label{pb1} For a fixed clustering $Y$ (and a given $\eta>0$),
\[ \frac{1}{2} \|Y\mu-XW\|_F^2 \to \min \]
under the constraint (\ref{constr3}) on $W$.
\end{problem}

\noindent Given the matrix of weights $W$, the second subproblem is the standard k-means on the projected data.

\begin{problem} \label{pb2} For a fixed projection matrix $W$, 
\[ \frac{1}{2} \|Y\mu-XW\|_F^2 \to \min \]
under the constraints (\ref{constr1})-(\ref{constr2}) on $Y$.
\end{problem}

\subsection{Exact gradient-projection splitting method}
\noindent To solve Problem~\ref{pb1},
we use a gradient-projection method. It belongs
to the class of splitting methods (\cite{BoydOpti,Smms05,Banf11,lions,Silvia,ML2011,BoydOptiProx}).
It is designed to solve minimization problems of the form
\begin{equation}\label{eq100}
  \vphi(W) \to \min,\quad W \in C,
\end{equation}
using separately the convexity properties of the function $\vphi$ on one hand, and
of the convex set $C$ on the other. We use
the following forward-backward scheme to generate a sequence of iterates:
\begin{eqnarray} \label{eq101}
  V_n & := & W_n - \gamma_n\nabla \vphi(W_n),\\
  W_{n+1} & := & P_C(V_n) + \veps_n, \label{eq102}
\end{eqnarray}
where $P_C$ denotes the projection on the convex set $C$ (a subset of some Euclidean
space). Under standard assumptions on the sequence of gradient steps $(\gamma_n)_n$,
and on the sequence of projection errors $(\veps_n)_n$, convergence holds (see,
\eg, \cite{Livre1}).

\begin{theorem} \label{th100}
Assume that (\ref{eq100}) has a solution.
Assume that $\vphi$ is convex, differentiable, and that $\nabla \vphi$ is
$\beta$-Lipschitz, $\beta>0$. Assume finally that $C$ is convex and that
\[ \sum_n |\veps_n| < \iy,\quad \inf_n \gamma_n > 0,\quad
   \sup_n \gamma_n < 2/\beta. \]
Then the sequence of iterates of the forward-backward scheme (\ref{eq101}-\ref{eq102})
converges, whatever the initialization.
If moreover $(\veps_n)_n = 0$ (exact projections), there exists a rank
$N$ and a positive constant $K$ such that, for $n \geq N$,
\begin{equation} \label{eq105}
  \vphi(W_n)-\inf_C \vphi \leq K/n.
\end{equation}
\end{theorem}

\noindent In our case, $\nabla \vphi$ is Lipschitz since it is affine,
\begin{equation}
 \nabla \vphi (W) = X^{T}(XW-Y\mu), 
\end{equation}
and we recall the estimation of its best Lipschitz constant.

\begin{lemma} \label{lmm1} Let $A$ be a $d \times d$ real matrix, acting linearly on the set of
$d \times k$
real matrices by left multiplication, $W \mapsto AW$. Then, its norm as a linear
operator on this
set endowed with the Frobenius norm is equal to its largest singular value,
$\sigma_\mathrm{max}(A)$.
\end{lemma}

\noindent\emph{Proof.} The Frobenius norm is equal to 
the $\ell^2$ norm of the vectorized matrix,
\begin{equation}
 \|W\|_F = \| \left[ \begin{array}{c} W^1\\ \vdots\\ W^h \end{array} \right] \|_2,\quad
   \|AW\|_F = \| \left[ \begin{array}{c} AW^1\\ \vdots\\ AW^h \end{array} \right] \|_2, 
\end{equation}
where $W^1,\dots,W^h$ denote the $h$ column vectors of the $d \times h$ matrix $W$.
Accordingly, the operator norm
is equal to the largest singular value of the $kd \times kd$ block-diagonal matrix
whose diagonal is made of $k$ matrix $A$ blocks.
Such a matrix readily has the same largest singular value as $A$.\hfill$\square$\\

\noindent As a byproduct of Theorem~\ref{th100}, we get

\begin{corollary} \label{cor1}
For any fixed step $\gamma \in (0,2/\sigma^2_\mathrm{max}(X))$,
the forward-backward scheme applied to the Problem~\ref{pb1} with an exact
projection on $\ell^1$ balls converges with a linear rate towards a solution,
and the estimate (\ref{eq105}) holds.
\end{corollary}

\noindent\emph{Proof.} The $\ell^1$ ball being compact, existence holds. So does
convergence, provided the condition of the step lengths is fulfilled. Now, according
to the previous lemma, the best Lipschitz constant of the gradient of $\varphi$ is
$\sigma_\mathrm{max}(X^{T} X)=\sigma^2_\mathrm{max}(X)$,
hence the result.\hfill$\square$\\

\begin{algorithm}
\begin{algorithmic}
\STATE \textbf{Input:} $X,Y,\mu,W_0,N,\gamma,\eta$
	\STATE $W \leftarrow W_0$
	\FOR{$n = 0,\dots,N$}
		\STATE $V \leftarrow W-\gamma X^{T}(XW-Y\mu)$
		\STATE $W \leftarrow P^1_{\eta}(V)$
	\ENDFOR
	\STATE \textbf{Output:} $W$
\end{algorithmic}
\caption{Exact gradient-projection algorithm}
\label{algo1}
\end{algorithm}

\noindent\textbf{Exact projection.}
In Algorithm~\ref{algo1}, we denote by $P^1_\eta(W)$ the
(reshaped as a $d \times \bar{d}$ matrix) projection of the vectorized matrix $W(:)$.
An important asset of the method is that it takes advantage of the availability of
efficient methods (\cite{condat,duchi}) to compute the $\ell^1$ projection.
For $\eta>0$, denote
$B^1(0,\eta)$ the closed $\ell_1$ ball of radius $\eta$ in the space $\RR^{d\times\bar{d}}$
centered at the origin, and
$\Delta_\eta$ the simplex $\{w \in \RR^{d\times\bar{d}}\ |\ w_1+\cdots+w_{d\bar{d}}=1,\ %
w_1 \geq 0,\dots,w_{d\bar{d}} \geq 0\}$. Let $w \in \RR^{d\times\bar{d}}$, and let $v$
denote the
projection on $\Delta_\eta$ of $(|w_1|,\dots,|w_{d\bar{d}}|)$. It is well known that
the projection of $w$ on $B^1(0,\eta)$ is
\begin{equation}
 (\epsilon_1(v_1),\dots,\epsilon_{kd}(v_{d\bar{d}})),\quad \epsilon_j := \sign(w_j),\quad
   j=1,\dots,d\bar{d}, 
\end{equation}
and the fast method described in (\cite{condat}) is used to compute $v$ with
complexity $O(d\times \bar{d})$.\\

\noindent\textbf{Fista implementation.}
A constant step
of suitable size $\gamma$ is used in accordance with Corollary~\ref{cor1}.
In our setting, a useful normalization of the design matrix $X$ is
obtained replacing $X$ by $X/\sigma_\mathrm{max}(X)$. This sets the Lipschitz constant
in Theorem~\ref{th100} to one.
The $O(1/n)$ convergence rate of the algorithm can be speeded up to $O(1/n^2)$
using a FISTA step (\cite{fista}). In practice we use a modified version (\cite{FistaAC})
which ensures convergence of the iterates, see Algorithm~\ref{algo1f}.
Note that for any fixed step $\gamma \in (0,1/\sigma^2_\mathrm{max}(X))$,
the FISTA algorithm applied to Problem~\ref{pb1} with an exact
projection on $\ell^1$ balls converges with a quadratic rate towards a solution,
and the estimate (\ref{eq105}) holds.

\begin{algorithm}
\begin{algorithmic}
\STATE \textbf{Input:} $X,Y,\mu,W_0,N,\gamma,\eta$
	\STATE $W \leftarrow W_0$
	\STATE $t \leftarrow 1$
	\FOR{$n = 0,\dots,N$}
		\STATE $V \leftarrow W-\gamma X^{T}(XW-Y\mu)$
		\STATE $W_\mathrm{new} \leftarrow P^1_{\eta}(V)$
            \STATE $t_\mathrm{new} \leftarrow (n+5)/4$
		\STATE $\lambda \leftarrow 1+(t-1)/t_\mathrm{new}$
		\STATE $W \leftarrow (1-\lambda)W+\lambda W_\mathrm{new}$
            \STATE $t \leftarrow t_\mathrm{new}$
	\ENDFOR
	\STATE \textbf{Output:} $W$
\end{algorithmic}
\caption{Exact gradient-projection algorithm with FISTA}
\label{algo1f}
\end{algorithm}

\subsection{Clustering algorithm}
\noindent The resulting alternating minimization is described by
Algorithm~\ref{algo2}.
(One can readily replace the gradient-projection step by the FISTA version described in
Algorithm~\ref{algo1f}.)
Labels $Y$ are for instance initialized by spectral clustering on $X$, while the
k-means computation relies on standard methods such as k-means++ (\cite{kmeanspp}).\\

\begin{algorithm}
\begin{algorithmic}
\STATE \textbf{Input:} $X,Y_0,\mu_0,W_0,L,N,k,\gamma,\eta$
\STATE $Y \leftarrow Y_0$
\STATE $\mu \leftarrow \mu_0$
\STATE $W \leftarrow W_0$
\FOR{$l = 0,\dots,L$}  
  \FOR{$n = 0,\dots,N$}
    \STATE $V \leftarrow W-\gamma X^{T}(XW-Y\mu)$
    \STATE $W \leftarrow P^1_{\eta}(V)$
  \ENDFOR
 \STATE $Y \leftarrow \texttt{kmeans}(XW,k)$
\STATE $\mu \leftarrow \texttt{centroids}(Y,XW)$ 
\ENDFOR
\STATE \textbf{Output:} $Y,W$
\end{algorithmic}
\caption{Alternating minimization clustering.}
\label{algo2}
\end{algorithm}

\noindent\textbf{Convergence of the algorithm.}
Similarly to the approaches advocated
in (\cite{diffrac,kanade,Ding,witten2010}),
our method involves non-convex k-means optimization for which convergence towards
local minimizers only can be proved (\cite{Bottou_nips,selim-1984a}).
In practice, we use k-means++
with several replicates to improve each clustering step. We assume that the initial
guess for labels $Y$ and matrix of weights $W$ is such that the associated $k$
centroids are all different.
We note for further research that there
have been recent attempts to convexify k-means (see, \eg,
\cite{PECOK,condat-2017a,mixon-2017a,peng-2007a}).
As each step of the alternating
minimization scheme decreases the norm in (\ref{cost}), which is nonnegative, the following readily holds.

\begin{proposition}
The Frobenius norm $\|Y\mu-XW\|_F$ converges as the number of
iterates $L$ in Algorithm~\ref{algo2} goes to infinity.
\end{proposition}

\noindent This property is illustrated in the next section on biological data.
Further analysis of the convergence may build on recent results on proximal
regularizations of the Gauss-Seidel alternating scheme for non convex problems
(\cite{proxnl,Bolte2014}).\\

\noindent\textbf{Gene selection.}
Feature selection is based on the sparsity inducing $\ell^1$ constraint (\ref{constr3}).
The projection $P^1_\eta(W)$ aims at sparsifying the $W$ matrix so that the gene $j$ will be selected if $\|W(j,:)\|>0$. 
For a given constraint $\eta$, the practical stopping criterion of the alternating
minimization algorithm involves the evolution of the number of the selected genes. At the
higher level loop on the bound
$\eta$ itself, the evolution of accuracy versus $\eta$ is analyzed.
We also note that the extension to multi-label classification
is straightforward as it suffices to allow several unit values on each
line of the matrix $Y$ by relaxing constraint (\ref{constr1b}).



\section{Experimental evaluation on single cell RNA-seq clustering} \label{s3}

\subsection{Experimental settings}

\noindent We normalize the features and use the FISTA implementation with constant step $\gamma = 1$  in accordance with Corollary~\ref{cor1}, and we set $\bar{d} = k+4$.
 Methods based on k-means provide different labels depending on the initial
conditions, thus we select the best result over $40$ replicates of k-means++ (\cite{kmeanspp}).
The problem
of estimating the number of clusters is out of the range of this study,
and we refer to the popular GAP method (\cite{kest}).
We compare the labels obtained from our clustering with the true labels to compute the
clustering accuracy.
We also report the popular \emph{Adjusted Rank Index}
(ARI) (\cite{ari2}) and \emph{Normalized Mutual Information} (NMI) criteria.
Processing times are obtained on a computer using an i7 processor (2.5 Ghz).\\ 
We compare our method with PCA k-means,
spectral clustering (\cite{Spectral}), SIMLR (Single-cell Interpretation via Multikernel Learning) (\cite{simlr, BachKernels}) and Sparcl (Sparse k-means clustering) (\cite{witten2010}). The first two methods (PCA k-means and spectral clustering) are standard and easily tested, while we have used the \textsl{R} software package \texttt{Sparcl} provided by (\cite{witten2010}) for Sparcl method. And we refer for SIMLR to the codes available online: See \href{https://github.com/BatzoglouLabSU/SIMLR/tree/SIMLR/MATLAB}{\texttt{https://github.com/BatzoglouLabSU/SIMLR/tree/SIMLR/MATLAB}}.

\subsection{Application to computational biology: Synthetic datasets}
\begin{figure}[!h]
\begin{center}
\includegraphics[width=0.6\linewidth,height=4cm]{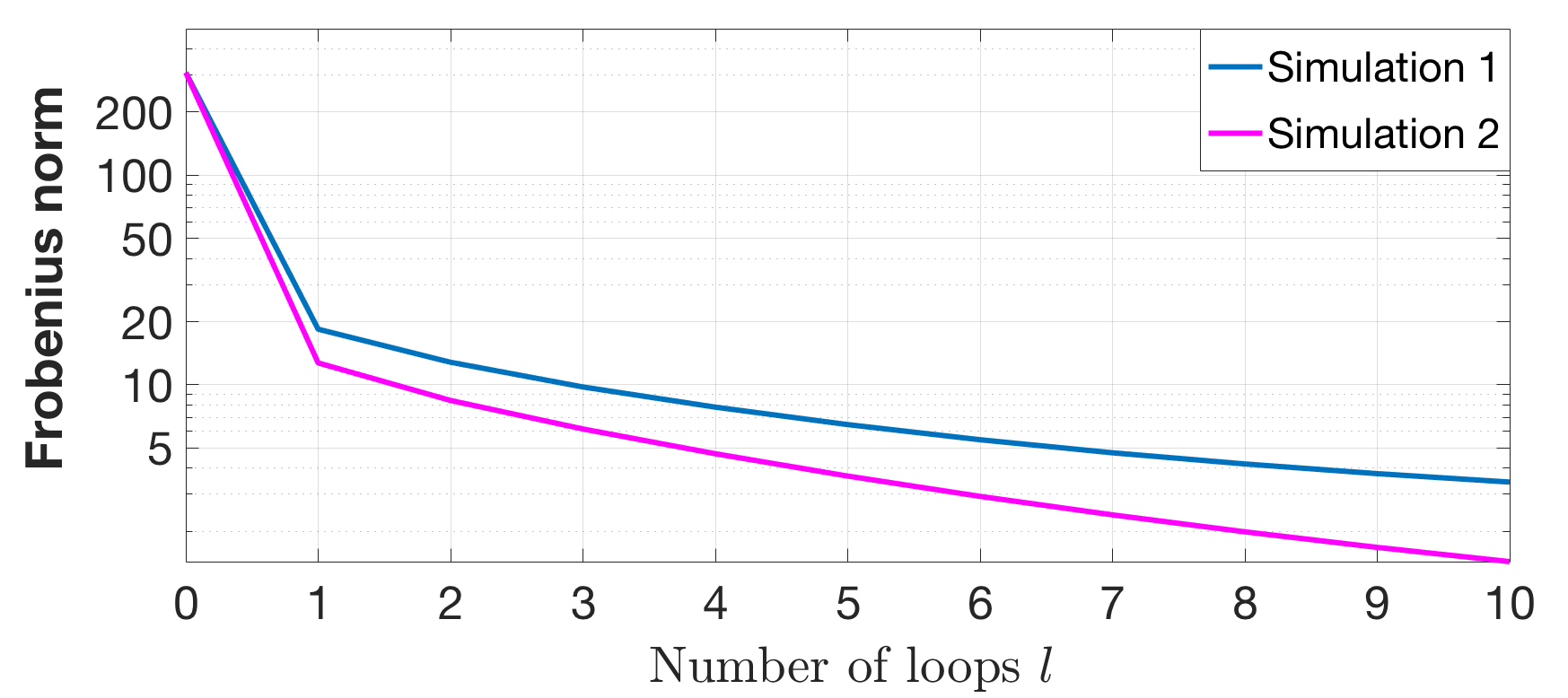}
\end{center}
\caption{\small Decay of the Frobenius norm for the two synthetic datasets versus the number of loops of the alternating minimization scheme 
emphasizes the fast and smooth convergence of our algorithm.}
\label{fig:norm}
\end{figure}

\begin{figure}[!h]
\includegraphics[width=0.5\linewidth,height=4cm]{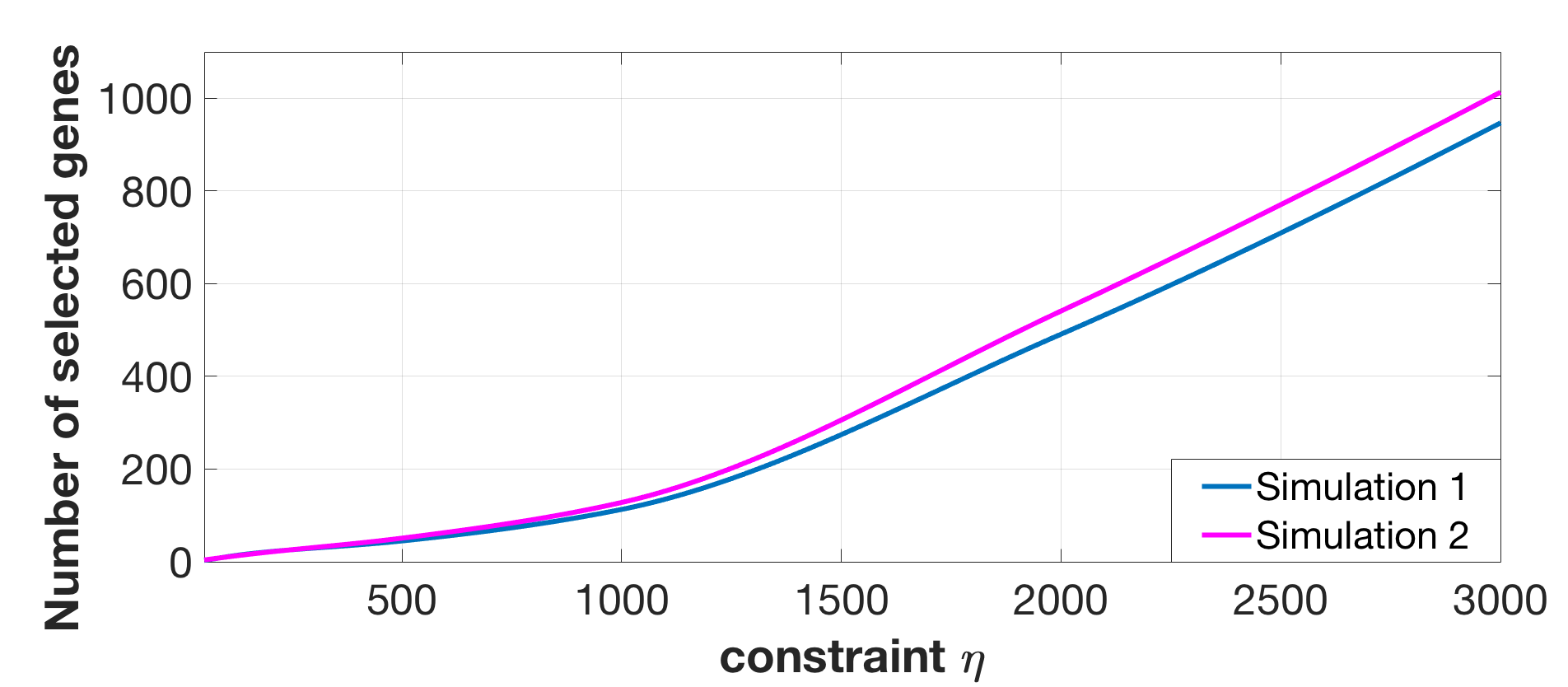}
\includegraphics[width=0.5\linewidth,height=4cm]{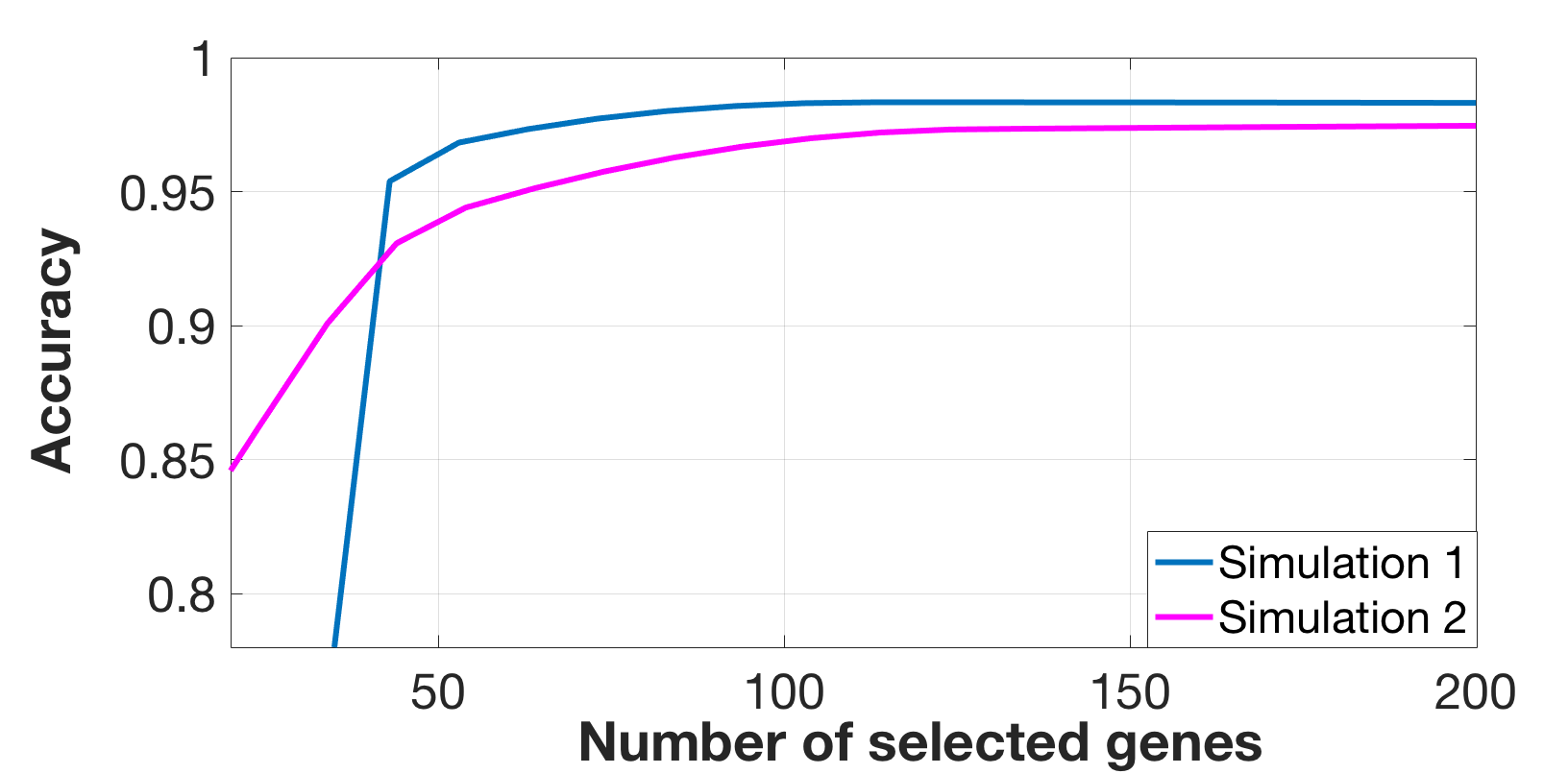}
\caption{\small Left: The evolution of the number of selected genes versus the constraint is a smooth monotonous function. The bound $\eta$ for the $\ell^1$ constraint is thus easily tuned.
Right: Accuracy versus number of genes. These results show that a minimum number of genes is required to get the best possible clustering accuracy.}
\label{fig:norm2}
\end{figure}

The simulation software was downloaded from \href{https://github.com/DeprezM/SCsim}{\texttt{https://github.com/DeprezM/SCsim}}. We use default parameters.
 The decay of the Frobenius norm (\ref{cost}) is portrayed Figure~\ref{fig:norm}, while the evolution of the number of selected genes vs.\ the sparsity constraint ($\ell^1$ constraint) or the accuracy is shown Figure~\ref{fig:norm2}. Both graphs illustrate the good properties of our method in terms of convergence, feature selection (a plateau in accuracy is reached as soon as the number of selected genes is large enough). As is clear from Tables~\ref{tab:simu1} and \ref{tab:simu2}, k-sparse behaves better than any of the four other methods on synthetic data. We also provide \texttt{tsne} (\cite{tsne}) for a 2D visual evaluation of each method (see Figure~\ref{fig:clusters}). The results, quite comparable for SIMLR and k-sparse, provide a clear confirmation of those in Tables~\ref{tab:simu1} and \ref{tab:simu2} .

%
%
\begin{table}[!h]
\begin{center}
\caption{\small Comparison between methods for the synthetic dataset 1 (4 clusters, 600 cells, 5,000 genes). For $\eta = 1000$ k-sparse selected $113$ genes (see Figure~\ref{fig:norm2}. Left.) and outperforms others methods in terms of accuracy, ARI and NMI.}
\label{tab:simu1}
\begin{tabular}{|c|c|c|c|c|c|}
\hline
\textbf{\textcolor{blue}{Simulation 1}} & \textcolor{blue}{PCA} & \textcolor{blue}{Spectral} & \textcolor{blue}{SIMLR} & \textcolor{blue}{k-sparse} \\
\hline
\textcolor{blue}{Accuracy} ($\% $) & 62.33 & 74.00 & 97.33 &  \textbf{98.33}   \\
\hline
\textcolor{blue}{ARI} ($\% $) & 37.21 & 56.43 & 93.77 &  \textbf{95.27}  \\
\hline
\textcolor{blue}{NMI}  & 0.50 & 0.63 & 0.89 &  \textbf{0.92}  \\
\hline
\textcolor{blue}{Time} ($s $) & \textbf{0.36} & 0.48 & 10.04   & 13.73 \\
\hline
\end{tabular}
\end{center}
\end{table}
\begin{table}[!h]
\begin{center}
\caption{\small Comparison between methods for the synthetic dataset 2 (4 clusters, 600 cells, 10,000 genes). For $\eta = 3000$ k-sparse selected $1,089$ genes (see Figure~\ref{fig:norm2}. Left.) and outperforms others methods in terms of accuracy, ARI and NMI.}
\label{tab:simu2}
\begin{tabular}{|c|c|c|c|c|c|}
\hline
\textbf{\textcolor{blue}{Simulation 2}} & \textcolor{blue}{PCA} & \textcolor{blue}{Spectral} & \textcolor{blue}{SIMLR} & \textcolor{blue}{k-sparse} \\
\hline
\textcolor{blue}{Accuracy} ($\% $) & 61.33 & 74.50 & 97.50 &  \textbf{97.83}   \\
\hline
\textcolor{blue}{ARI} ($\% $) & 34.75 & 57.36 & 93.26 &  \textbf{94.03}  \\
\hline
\textcolor{blue}{NMI}  & 0.49 & 0.60 & 0.90 &  \textbf{0.91}  \\
\hline
\textcolor{blue}{Time} ($s $) & \textbf{0.63} & 0.75 & 13.82   & 62.92 \\
\hline
\end{tabular}
\end{center}
\end{table}
\begin{figure*}[!h]
\begin{center}
\includegraphics[width=1\linewidth,height=6.5
cm]{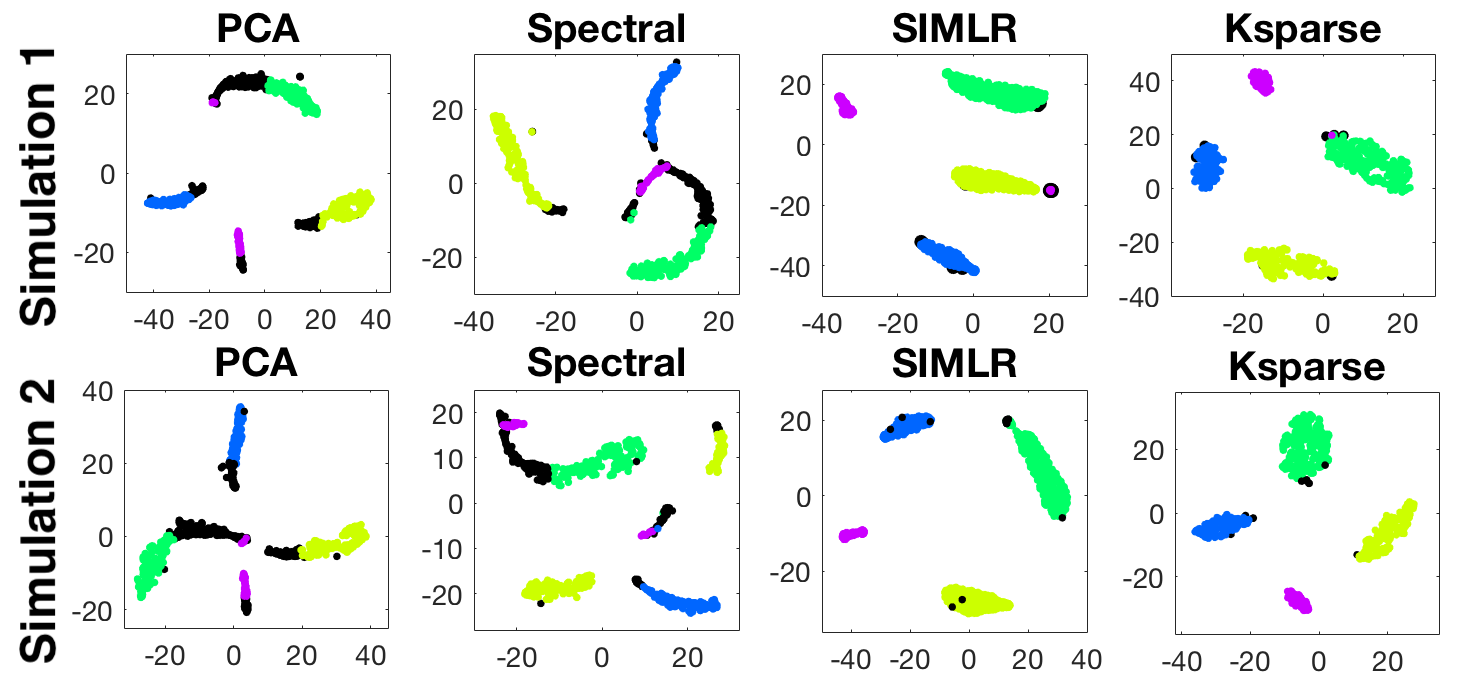}
\caption{\small Comparison of 2D visualization using \texttt{tsne}  (\cite{tsne}).
Each point represents a cell. Misclassified cells in black are reported for each method. This figure shows the nice small ball-shaped clusters computed by k-sparse and SIMLR methods.}
\label{fig:clusters}
\end{center}
\end{figure*}

\clearpage

\subsection{Application to computational biology: Single cell datasets}
\noindent Our algorithm can be
readily extended to multiclass clustering of high dimensional databases in
computational biology (single cell clustering, mass-spectrometric data...), pattern
recognition, combinatorial chemistry, social networks clustering, decision making,
\emph{etc.}
We provide an experimental evaluation on Single-cell sequencing dataset.
The new Single-cell technology has been elected "method of the year" in 2013 by {\em Nature Methods} (\cite{SingleCellMthdYear}).
The widespread use of such methods has enabled the publication of many datasets with ground truth cell type annotations (\cite{src}). Thus we compare algorithms on three of those public single-cell RNA-seq datasets:  Klein dataset (\cite{klein}), Zeisel dataset (\cite{zeisel})
and Usoskin (\cite{usoskin}) dataset.\\

\noindent \textbf{Klein scRNA-seq dataset. }
\cite{klein} characterized the transcriptome of 2,717
cells (\emph{Mouse Embryonic Stem Cells}, mESCs),
across four culture conditions (control and with 2, 4 or
7 days after leukemia inhibitory factor, LIF, withdrawal) using InDrop sequencing. Gene expression was quantified
with \emph{Unique Molecular Identifier} (UMI) counts (essentially tags that identify
individual molecules allowing removal of amplification bias). The raw UMI
counts and cells label were downloaded from
\href{http://hemberg-lab.github.io/scRNA.seq.datasets}{\tt hemberg-lab.github.io/scRNA.seq.datasets}.
After filtering out lowly expressed genes (10,322 genes remaining after removing genes that have less than 2 counts in 130 cells) and Count Per Million normalization (CPM) to reduce cell-to-cell variation in sequencing, we
report clustering into four cell sub-populations, corresponding to the four culture conditions.\\

\noindent \textbf{Zeisel scRNA-seq dataset. }
Zeisel et al. (\cite{src, zeisel}) collected 3,005 mouse cells from the primary somatosensory
cortex (S1) and the hippocampal CA1 region, using the Fluidigm C1 microfluidics cell
capture platform followed. Gene expression was quantified with UMI counts. The raw UMI counts and metadata (batch, sex,
labels) were downloaded from
\href{http://linnarssonlab.org/cortex}{\tt linnarssonlab.org/cortex}.
We applied low expressed gene filtering (7,364
remaining genes after removing genes that have less than 2 counts in 30 cells) and CPM normalization. We report clustering into the nine major classes identified in the study.\\

\noindent \textbf{Usoskin scRNA-seq dataset. }
Uzoskin et al. (\cite{usoskin}) collected 622 cells from the mouse dorsal root
ganglion, using a robotic cell-picking setup and sequenced with a 5' single-cell
tagged reverse transcription (STRT) method. Filtered (9,195 genes) and normalized data
(expressed as Reads Per Million) were downloaded with full sample annotations from
\href{http://linnarssonlab.org/drg}{\tt linnarssonlab.org/drg}.
We report clustering into four neuronal cell types.


\subsection{Comparison between methods }
\begin{figure}[!h]
\begin{center}
\includegraphics[width=0.6\linewidth,height=4cm]{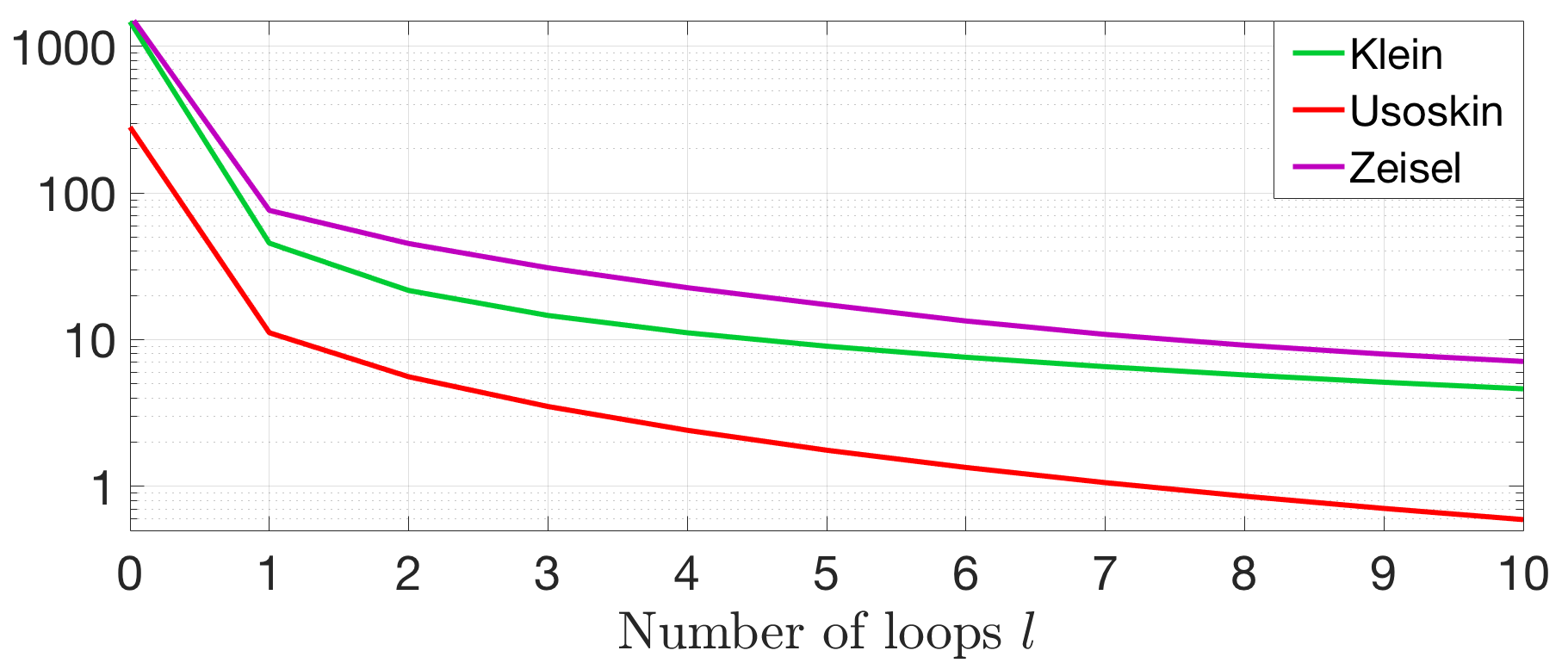}
\end{center}
\caption{\small Left: Decay of the Frobenius norm for the three  datasets versus the number of loops of the alternating minimization scheme emphasizes the fast and smooth convergence of our algorithm. 
}
\label{fig:frob100}
\end{figure}

\begin{figure}[!h]
\includegraphics[width=0.5\linewidth,height=4cm]{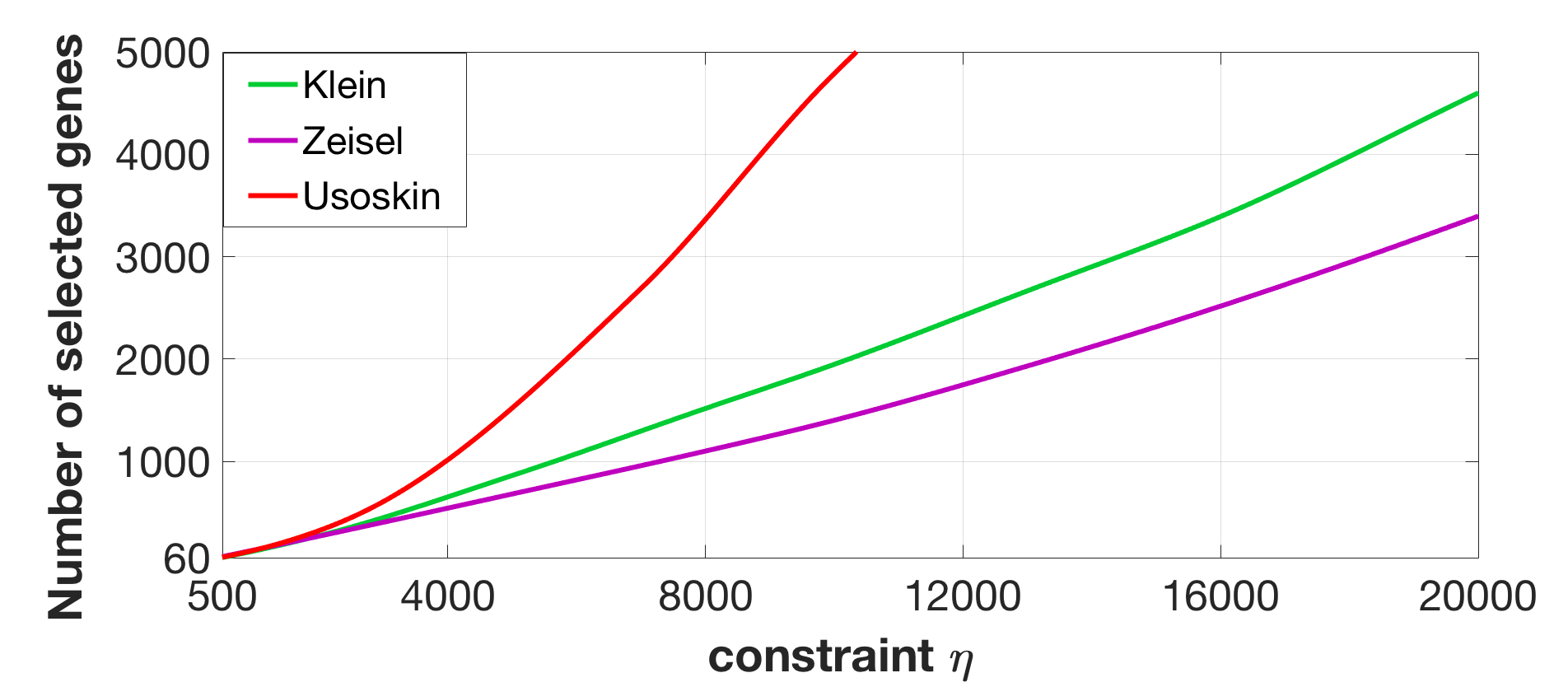}
\includegraphics[width=0.5\linewidth,height=4cm]{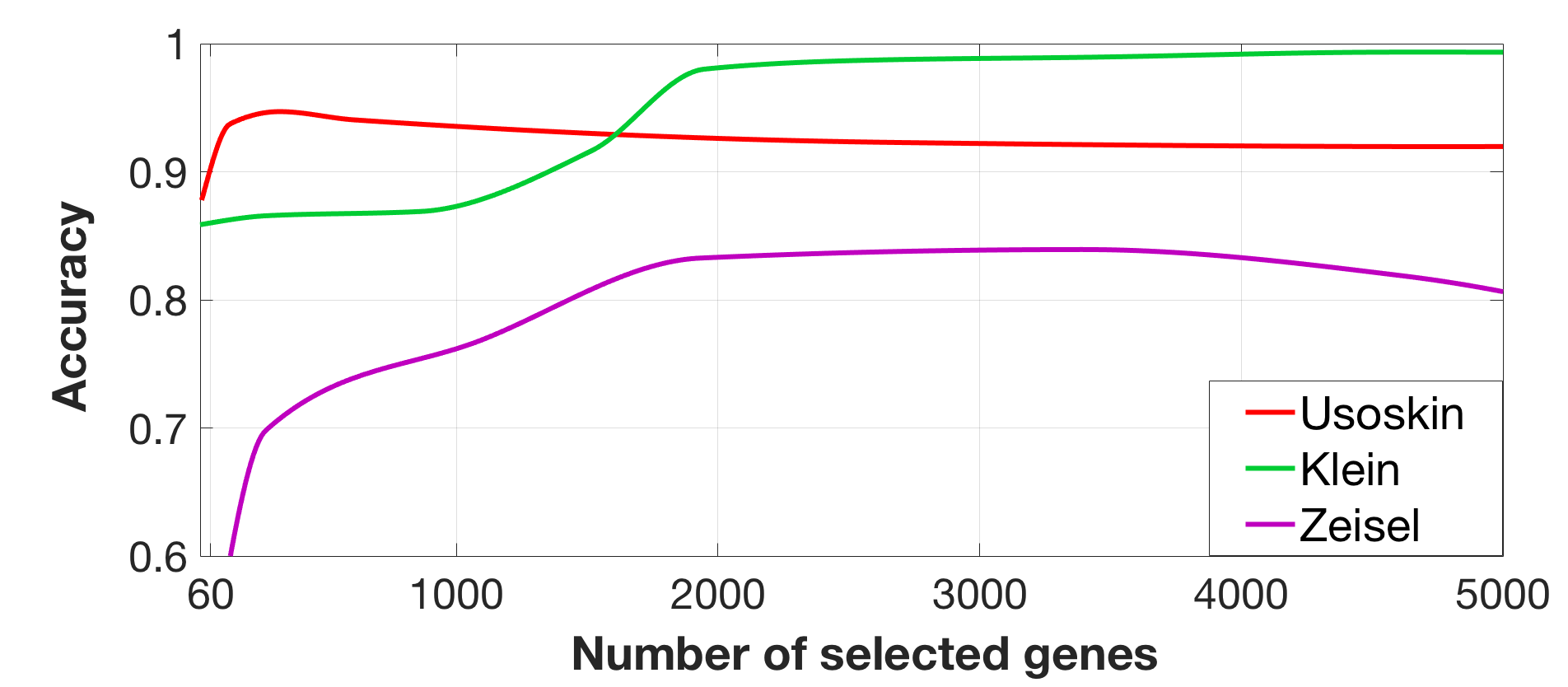}
\caption{\small Left: The evolution of the number of selected genes versus the constraint is a smooth monotonous function. In our constrained approach, parameter $\eta$ is directly connected to the number of genes. An estimate of this number is known by biologists. Thus we tune $\eta$ in order to obtain the desired number of genes. Right: We use Accuracy, ARI and NMI (which indeed uses ground-true labels) for evaluation only of our algorithm, and comparison with contenders. These results show that a minimum number of
genes is required to get the best possible clustering accuracy and  that accuracy is constant over a large plateau, making the tuning of the parameter very easy.} 
\label{fig:norm100}
\end{figure}

We provide accuracy, ARI, NMI and time processing for five different methods: PCA k-means,
spectral clustering (\cite{Spectral}), SIMLR (Single-cell Interpretation via
Multikernel Learning) (\cite{simlr}), Sparcl (Sparse k-means
clustering) (\cite{witten2010}), and our method k-sparse.
As in the previous section on synthetic data, we provide an evaluation of k-sparse on each of three bases (Klein, Usoskin and Zeisel) in terms of convergence (see Figure~\ref{fig:frob100}), feature selection and accuracy (Figure~\ref{fig:norm100}).
Our method significantly improves the results of Sparcl and SIMLR in terms of accuracy, ARI and NMI. For each of the three databases, k-sparse obtains the best results when compared to the four other methods, not only for accuracy but also for ARI and NMI (see Tables~\ref{tab:usoskin}, \ref{tab:klein} and \ref{tab:zeisel}. We note however that k-sparse, though faster than SIMLR in two cases out of three, has larger execution times than much less precise methods such as PCA k-means (for which very efficient codes exist). We provide again \texttt{tsne} (\cite{tsne})
for visual evaluation and comparison of the five methods (Figure~\ref{fig:clusters}).

\clearpage


\begin{table}[!h]
\begin{center}
\caption{\small Comparison between methods for Usoskin dataset (4 clusters, 622 cells, 9,195 genes). For $\eta = 8000$, k-sparse selected $3,325$ genes (see Figure~\ref{fig:norm100}. Left.) and outperforms others methods in terms of accuracy by $15 \%$.}
\label{tab:usoskin}
\begin{tabular}{|c|c|c|c|c|c|c|}
\hline
\textbf{\textcolor{blue}{Usoskin  dataset}} & \textcolor{blue}{PCA} & \textcolor{blue}{Spectral} & \textcolor{blue}{SIMLR}  & \textcolor{blue}{Sparcl}& \textcolor{blue}{k-sparse} \\
\hline
\textcolor{blue}{Accuracy} ($\% $) & 54.82 & 60.13 & 76.37 & 57.24 & \textbf{91.96}   \\
\hline
\textcolor{blue}{ARI} ($\% $) & 22.33 & 26.46 & 67.19 & 31.30 & \textbf{85.85}  \\
\hline
\textcolor{blue}{NMI}  & 0.29 & 0.33 & 0.75 & 0.39 & \textbf{0.83}  \\
\hline
\textcolor{blue}{Time} ($s $) & 1.06 & \textbf{0.91} & 15.67  & 1,830 & 57.07 \\
\hline
\end{tabular}
\end{center}
\end{table}

\begin{table}[!h]
\begin{center}
\caption{\small Comparison between methods for Klein dataset (4 clusters, 2,717 cells, 10,322 genes). For $\eta = 20000$, k-sparse selected $4,599$ genes (see Figure~\ref{fig:norm100}. Left.) and has an accuracy close to $100\%$. SIMLR has similar performances (accuracy, ARI and NMI) than k-sparse (which is $5$ times faster than SIMLR).}
\label{tab:klein}
\begin{tabular}{|c|c|c|c|c|c|c|}
\hline
\textbf{\textcolor{blue}{Klein dataset}} & \textcolor{blue}{PCA} &\textcolor{blue}{Spectral}  & \textcolor{blue}{SIMLR}  & \textcolor{blue}{Sparcl}& \textcolor{blue}{k-sparse} \\
\hline
\textcolor{blue}{Accuracy} ($\% $) & 68.50 & 63.31 & 99.12 & 65.11 & \textbf{99.33}  \\
\hline
\textcolor{blue}{ARI} ($\% $) & 44.82 & 38.91& 98.34 & 45.11 & \textbf{98.77}  \\
\hline
\textcolor{blue}{NMI}  & 0.55 & 0.54 & 0.96 & 0.56 & \textbf{0.97}  \\
\hline
\textcolor{blue}{Time} ($s $) & \textbf{10.91} & 20.81 & 511 &  30,384 & 97.10 \\
\hline
\end{tabular}
\end{center}
\end{table}


\begin{table}[!h]
\begin{center}
\caption{\small Comparison between methods for Zeisel dataset (9 clusters, 3,005 cells, 7,364 genes). For $\eta = 16000$, k-sparse selected $2,497$ genes (see Figure~\ref{fig:norm100}. Left.) and outperforms others methods in terms of accuracy by $11 \%$. K-sparse is $6$ times faster than SIMLR.} 

\label{tab:zeisel}
\begin{tabular}{|c|c|c|c|c|c|c|}
\hline
\textbf{\textcolor{blue}{Zeisel dataset}} & \textcolor{blue}{PCA} &\textcolor{blue}{Spectral}  & \textcolor{blue}{SIMLR}  & \textcolor{blue}{Sparcl}  & \textcolor{blue}{k-sparse} \\
\hline
\textcolor{blue}{Accuracy} ($\% $) &39.60 & 59.30 & 71.85 & 65.23 & \textbf{83.26} \\
\hline
\textcolor{blue}{ARI} ($\% $) & 34.67 & 50.55 &  64.8 & 59.06 & \textbf{75.06}  \\
\hline
\textcolor{blue}{NMI}  & 0.54 & 0.68 & 0.75 & 0.69 & \textbf{0.77}  \\
\hline
\textcolor{blue}{Time} ($s $) & \textbf{11} & 23 & 464  & 28,980 &  71.60 \\
\hline
\end{tabular}
\end{center}
\end{table}

\subsection{Scalability}
K-sparse converges within around $L=10$ loops.
The complexity of the inner iteration of k-sparse is $O(d\times \bar{d}\times
d_n(\eta))$ for the gradient part (sparse matrix multiplication $X^{T}XW$),
plus $O(d\times\bar{d})$ for the projection part, where $d_n(\eta)$ is the average number of
nonzero entries of the sparse matrix $W$. This number depends on the sharpness of the
$\ell^1$ constraint (\ref{constr3}) defined by $\eta$, and on the iteration $n$. (As
$n$ ranges from $0$ to $N$, sparsity is increased as illustrated by the numerical
simulations.) The number of genes decreases rapidly with the iterates which allows to use sparse computing.
One must then add the cost of k-means, that is expected to be $O(m\times\bar{d})$ in average.
This allows k-sparse to scale up to large or very large databases.
In contrast, optimizing the values of the Lagrangian parameter using permutations
Sparcl is computationally expensive, with complexity $O(m^2 \times d)$. Naive implementation of Kernel methods SIMLR results in $O(m^2)$ complexity. The computational cost can be  
reduced to $O(p^2 \times m)$ ($p$ is the low rank) using low rank kernel matrix approximation (\cite{Bach13}). The computational cost is improved (see Table~\ref{tab:syntheticTime}) while the performance (ARI) drop significantly (see Table~\ref{tab:synthetic}) when using low rank kernel matrix approximation in Large SIMLR (\href{https://github.com/BatzoglouLabSU/SIMLR/tree/SIMLR/MATLAB}{\texttt{https://github.com/BatzoglouLabSU/SIMLR/tree/SIMLR/MATLAB}}).

\begin{table}[!h]
\begin{center}
\caption{\small Comparison between SIMLR, Large SIMLR and k-sparse in terms of ARI ($\% $) on large datasets. K-sparse outperforms Large SIMLR by $36\%$ on Klein dataset and $20\%$ on Zeisel dataset in terms of ARI.}
\label{tab:synthetic}
\begin{tabular}{|c|c|c|c|c|c|c|c|}
\hline
\textbf{\textcolor{blue}{Methods }} & \textcolor{blue}{SIMLR}  & \textcolor{blue}{Large SIMLR}  &\textcolor{blue}{k-sparse} \\
\hline
\textcolor{blue}{Klein}  (2,717 cells, $10,322$ genes, $k=4$)  &  98.34   &  61.49   & \textbf{98.77}  \\
\hline
\textcolor{blue}{Zeisel}   (3,005 cells, $7,364$ genes, $k=9$)  & 64.8 &56.39 & \textbf{75.06} \\
\hline
\end{tabular}
\end{center}
\end{table}

\begin{table}[!h]
\begin{center}
\caption{\small Comparison between SIMLR, Large SIMLR and k-sparse in terms of time ($s$) on large datasets. K-sparse is $8$ times faster on Klein dataset and $10$ times faster on Zeisel dataset  than SIMLR. Large SIMLR is faster than k-sparse but Table \ref{tab:synthetic} shows that the clusters performed by Large SIMLR are not similar to real clusters.} 
\label{tab:syntheticTime}
\begin{tabular}{|c|c|c|c|c|c|c|c|}
\hline
\textbf{\textcolor{blue}{Methods }}  & \textcolor{blue}{SIMLR}  &\textcolor{blue}{Large SIMLR}  &\textcolor{blue}{k-sparse} \\
\hline
\textcolor{blue}{Klein}  (2,717 cells, $10,322$ genes, $k=4$) &   511 & \textbf{8.64}  & 97.10  \\
\hline
\textcolor{blue}{Zeisel}  (3,005 cells, $7,364$ genes, $k=9$) &  464  & \textbf{8.19}  & 71.60 \\
\hline
\end{tabular}
\end{center}
\end{table}



\begin{figure*}[!h]
\begin{center}
\includegraphics[width=1\linewidth,height=18
cm]{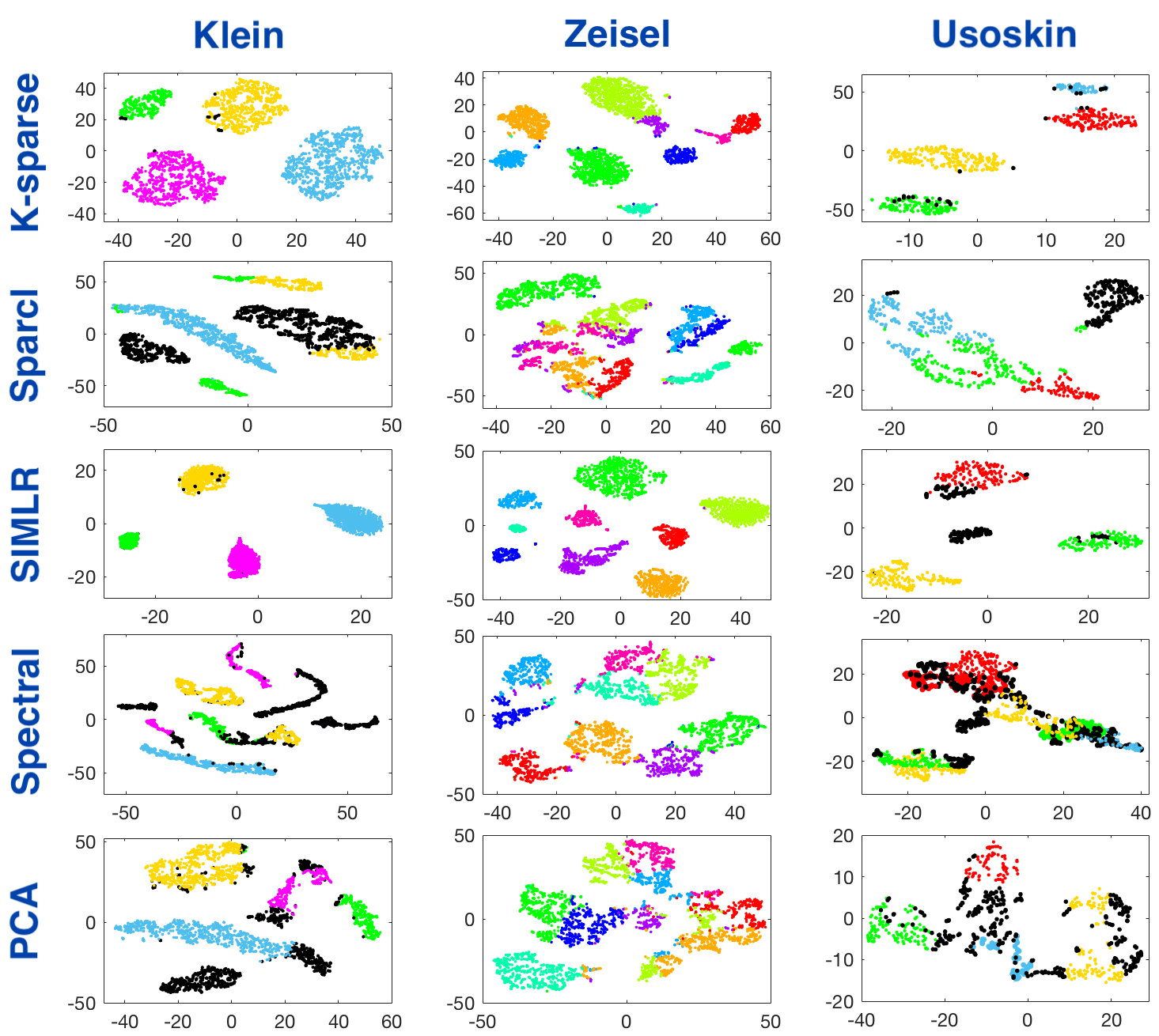}
\caption{\small Comparison of 2D visualization using \texttt{tsne}  (\cite{tsne}).
Each point represents a cell. Misclassified cells in black are reported for two
datasets: Klein and Usoskin. k-sparse significantly  improves visually the results of
Sparcl and SIMLR (note that SIMLR fails to discover one class on Usoskin). This figure
shows the nice small ball-shaped clusters computed by k-sparse and SIMLR methods.}
\label{fig:clusters}
\end{center}
\end{figure*}

\section{Conclusion} \label{s4}
\noindent In this paper,
we focus on unsupervised classification. We provide a new efficient
algorithm based on alternating minimization that achieves feature selection by
introducing an $\ell^1$ constraint in the gradient-projection step. This step, of
splitting type, uses an exact projection on the $\ell^1$ ball to promote sparsity, and is
alternated with k-means.
Convergence of the  projection-gradient method is established, and
each iterative step of our algorithm necessarily lowers the
cost.
Experiments on single-cell RNA-seq dataset in Section~\ref{s3}
demonstrate that our method is very promising compared to other algorithms in the field. 
Ongoing developments deal with the application of k-sparse to very large datasets. 

\small
\bibliography{JMLR_18}

\end{document}